\documentclass[runningheads]{llncs}
\usepackage{graphicx}

\usepackage[backend=biber,minnames=2, maxnames=2, sorting=none]{biblatex}
\newcommand\blfootnote[1]{%
  \begingroup
  \renewcommand\thefootnote{}\footnote{#1}%
  \addtocounter{footnote}{-1}%
  \endgroup
}

\graphicspath{ {./images/} }

\begin{document} 
 
\title{Transformer Ensembles for Sexism Detection}
%
%
\author{
Lily Davies\inst{1}
\and
Marta Baldracchi\inst{1}
\and
Carlo Alessandro Borella\inst{1} 
\and
Konstantinos Perifanos\inst{1,2}\orcidID{0000-0002-8352-82943}
}

\authorrunning{EXIST2021-codec.ai}
%
\institute{codec.ai, London, UK  \and
National and Kapodestrian University of Athens, Greece\\
}
\maketitle              

\begin{abstract}

This document presents in detail the work done for the sexism detection task at EXIST2021 workshop. Our methodology is built on ensembles of Transformer-based models which are trained on different background and corpora and fine-tuned on the provided dataset from the EXIST2021 workshop. We report accuracy of 0.767 for the binary classification task (task1), and f1 score 0.766,  and for the multi-class task (task2) accuracy 0.623 and  f1-score 0.535.

\end{abstract}

\blfootnote{Copyright \textcopyright\ 2021 for this paper by its authors. Use permitted under Creative Commons License Attribution 4.0 International (CC BY 4.0). IberLEF 2021, September 2021, Málaga, Spain.}

\section{Introduction}

EXIST workshop (sEXism Identification in Social neTworks) is the first shared task at IberLEF 2021 \cite{iberlef_2021_proceedings}. The topic of this particular workshop is to build classifiers for sexism detection.  

\subsection{Dataset and Tasks}

As part of the workshop, an annotated dataset \cite{EXIST2021} has been provided, which consists of sexist expressions, both in English and Spanish, commonly used on social media. The source of these texts are Twitter and Gab social media platforms. 

The training dataset consists of 6977 texts, 3436 in English and 3541 in Spanish. In the training set, all of the texts have Twitter as their source. The training dataset has been labelled with two separate label sets,  which are essentially the workshop tasks: first, a higher level binary annotation per text, indicating whether the particular text is sexist or not, and a second layer of annotation per text, where if a particular text is identified as sexist, it is also assigned one of the following labels : \{IDEOLOGICAL AND INEQUALITY, STEREOTYPING AND DOMINANCE, OBJECTIFICATION, SEXUAL VIOLENCE, MISOGYNY AND NON-SEXUAL VIOLENCE\}. 

For the first task, the dataset contains 3377 texts labelled as sexist and 3600 labelled as non-sexist, so it is rather balanced. For the second task, the distribution of the tweets labelled as sexist in their subcategories is as shown in Table \ref{tab1}:  

\begin{table}
\centering
\caption{Number of labels per class, training dataset.}\label{tab1}
\begin{tabular}{|l|l|}
\hline
\bfseries{Label} &  \bfseries{Count of examples} \\
\hline
objectification &  500 \\
sexual-violence &  517\\
misogyny-non-sexual-violence & 685 \\
stereotyping-dominance &  809 \\
ideological-inequality & 866 \\
non-sexist &   3600 \\
\hline
\end{tabular}
\end{table}

The test dataset consists of 4368 texts, 2208 in English and 2160 in Spanish. In this dataset 3386 of the texts are sourced from Twitter and 982 from the social network Gab.

\section{System Architecture}

We combine two major approaches in our design: fine-tuning separate BERT \cite{devlin-etal-2019-bert} models trained in Spanish and English; and building ensembles of models per language.

\subsection{Task1}
We fine-tune six separate models starting from different weight configurations, three for English and three for Spanish.

During testing, we report the majority vote of the ensembles, e.g. if more than one of the classifiers agree on a prediction, we report that prediction as the final decision. Our work is following the reasoning and the results of \cite{allen-zhu2020towards}, where several models are trained in the same dataset with the same loss function but starting from a different random weight initialisation. The majority vote of the model ensemble then tends to perform better by each standalone model.   

In the training process, we fine-tune the BERT models \cite{attention} on the training data, using an 80-20 training-test split for each model, per language, where the languages available in the dataset are English and Spanish. This choice follows an empirical observation that language-specific pretrained BERT architectures tend to capture better the language subtleties for the tasks posed and tend to perform better in classification benchmarks.  

For English texts we use the pre-trained model from \cite{mishra-etal-2020-multilingual}, available in the Huggingface transformers library \cite{wolf2020huggingfaces}. For Spanish texts we fine-tune the BETO pre-trained model \cite{CaneteCFP2020}. 

We use PyTorch \cite{paszke2017automatic} for the implementation. 

Before feeding the texts to the classifiers, we pre-process them by replacing mentions with the \_\_mention\_\_ token and URLs by the  \_\_URL\_\_ token. While theoretically user mentions and handles can implicitly capture social graph structure and potentially increase the classifier performance \cite{perifanos_idiolect}, we choose to rely only on textual information and discard social graph or source for the tasks in question. 

We first filter the input training set based on the language indicator and then seed three neural networks with different initial random weights. We train the networks for 10 epochs and select the one with the best accuracy as the final model per training. 
In testing, we feed the same text to all three neural networks and report the majority vote of the classifiers as the final classification. The final output will be classified as sexist if at least two of the three classifiers report it as sexist and similarly, a text will be classified as non-sexist if at least two out of three classifiers report it as non-sexist. Based on our experiments this tends to give a $\sim2\%$ increase in the overall reported accuracy. We use Adam optimiser for the training \cite{kingma2017adam}.

\subsection{Task2}

For the second task, we train a classifier to distinguish between sexist tweets. That is, we only keep the texts in the training dataset that have been flagged as sexist and we train a multi-class model for these texts. 

Again, we train different models for Spanish and English language texts. Since we have the language labels of the texts there is no need to attempt to determine the language of a text; however, in the general case it would be straightforward to add one more step in the pipeline for language detection using an off the shelf language detection library, such as langdetect (https://pypi.org/project/langdetect/).

In testing time, we first apply the classifier of task1 to detect whether a tweet is sexist or not. If the text is labelled as non-sexist we report this label for task2 as well. If the task1 model reports sexist as the label, we feed this tweet to the second classifier to obtain a prediction for the sexism categorisation label as described above.

\section{Results and Discussion}

For the first task, we achieve accuracy of 0.767, with f-score (macro) of 0.766 (10th placement in the ranking). For the second task, accuracy is at 0.623 and f1-score at 0.535 (19th placement in the ranking).

Breaking down the final results by language,  for task1 the accuracy in English texts is at 0.7445 and for Spanish texts at 0.789. For task2, the accuracy in English texts is 0.583 with f1-score 0.493, whereas in Spanish texts accuracy is at 0.664 and f1-score at 0.575. 

Overall, the system tends to perform better in Spanish, and this is most probably due to the fact that the underlying BERT model for Spanish (BETO) is trained exclusively in Spanish texts.

Whereas we are using ensembles in the first task, we decided not to adopt this strategy for the multiclass classification for task2, as it would have increased significantly training and testing time as well as computational cost. Instead, we are using the output of the first task and we only train one classifier per language for the second task. It is worth noting here that the observation from \cite{allen-zhu2020towards} is validated in this case, as the ensemble achieves up to 2\% higher accuracy than standalone models.

\subsection{Analysis}

It is interesting to see that our models fail to correctly identify texts that have been labelled as non-sexist due to the presence of words that commonly appear in sexist environments. It also fails to correctly categorise sexist tweets that appear in the same context with non-sexist words (for example, the word \textit{friend}), short texts or even sometimes subtle or contextual use of sexists language.

\begin{figure}
\includegraphics[width = 6cm]{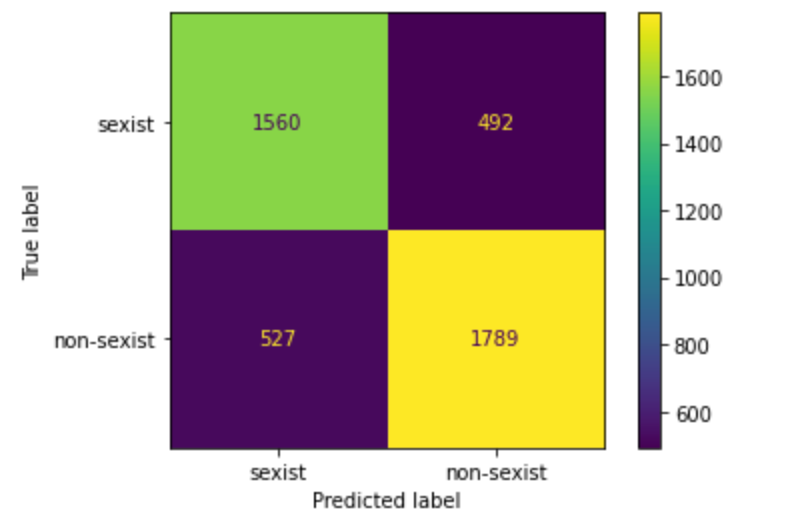}
\includegraphics[width = 6cm]{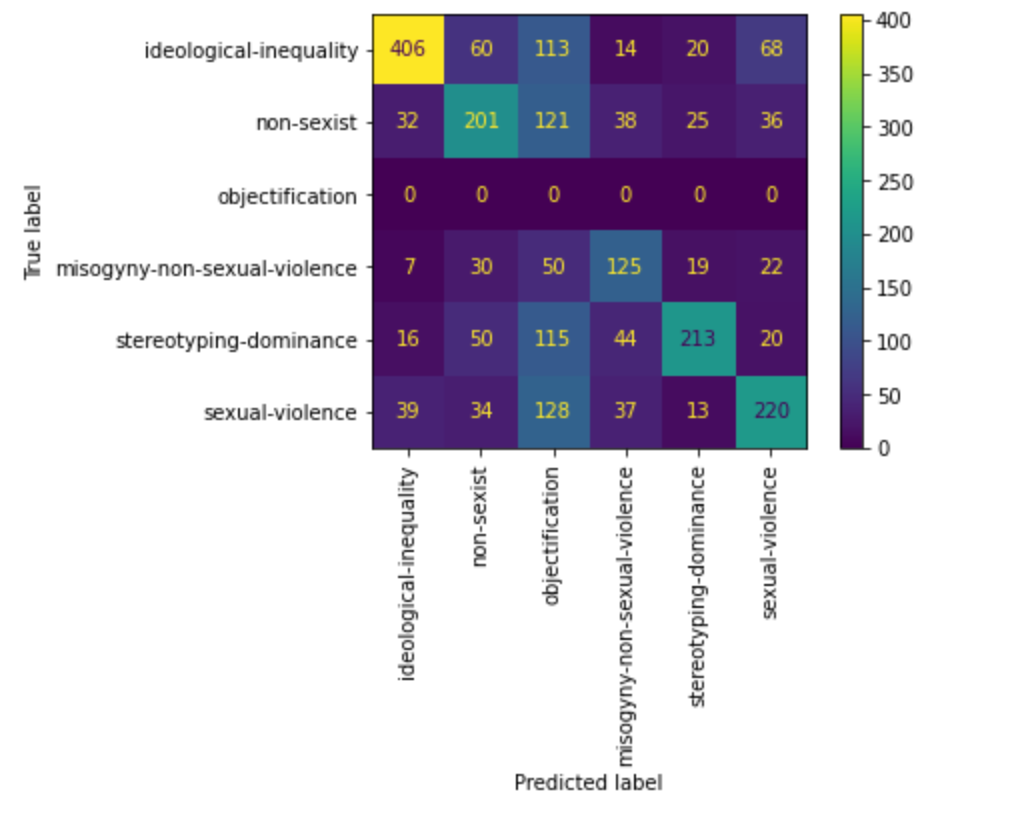}
\caption{Confusion matrices}
\label{Confusion_matrix}
\end{figure}

For task2, the predictions are inheriting the error from the task1 classifier, e.g. falsely reporting sexist tweets, which accumulates with the error of the second model. The confusion matrices for the 2 tasks are shown in figure \ref{Confusion_matrix}.

\subsection{Improvements and future research}

One question we aim to investigate is what is the optimal number of models in the ensemble from a classification accuracy perspective. This is a more general question and direction to investigate for Transformer based models for text classification.

Additionally,  most recently it has been shown that Convolutional Neural Networks can over-perform Transform-based architectures in Natural Language Processing tasks \cite{tay2021pretrained}. The use and fine-tuning of pre-trained convolutions in the domain of abusive and toxic speech would be an interesting direction to investigate.

%
%
%


%
\printbibliography

\end{document}